\documentclass[5p,times]{elsarticle}
\usepackage[T1]{fontenc}
\makeatletter
\IfFileExists{elsevier-logo-5p.pdf}{}{}
\IfFileExists{MFGLET-CRC.pdf}{}{}
\makeatother   % correct braces/underscores in \texttt trace listings

\journal{the 55th SME North American Manufacturing Research Conference (NAMRC 55, 2027), Manufacturing Letters}

\usepackage{amssymb}
\usepackage{amsmath}
\usepackage{booktabs}
\usepackage{multirow}
\usepackage{makecell}
\usepackage{siunitx}
\usepackage{xcolor}
\usepackage{graphicx}
\usepackage{array}
\usepackage{url}
\usepackage[figuresright]{rotating}
\usepackage[bookmarks=false]{hyperref}
\hypersetup{colorlinks, linkcolor=blue, citecolor=blue, urlcolor=blue}
\usepackage[capitalize]{cleveref}
\crefformat{appendix}{#2#1#3}\Crefformat{appendix}{#2#1#3}
\crefmultiformat{appendix}{#2#1#3}{ and~#2#1#3}{, #2#1#3}{ and~#2#1#3}
\usepackage[protrusion=true,expansion=false]{microtype}
\usepackage{placeins}
\usepackage{dblfloatfix}
\biboptions{sort&compress}
\graphicspath{{./}{./figs/}}

\newcommand{\ans}[1]{\texttt{#1}}
\newcommand{\repourl}{\url{https://github.com/Bfujah/umw-agent-benchmark}}
\newcommand{\evalwindow}{July 2026}

\begin{document}
\begin{frontmatter}

% \dochead{55th SME North American Manufacturing Research Conference (NAMRC 55, 2027)}%

\title{Orchestration reliability and failure modes of tool-using LLM agents for
manufacturing decision support}

\author[a]{Olabode Ajenifujah}
\author[a]{Zhiqiao Dong}
\author[b]{Baris Erol}
\author[b]{Hassan Ghassemi-Armaki}
\author[b]{Alp Ozdemir}
\author[b]{Jorge F. Arinez}
\author[a]{Jeffrey A. Abell\corref{cor1}}
\ead{jaabell@umich.edu}
\author[a]{Chenhui Shao\corref{cor1}}
\ead{chshao@umich.edu}

\address[a]{Department of Mechanical Engineering, University of Michigan, Ann Arbor, MI 48109, United States}
\address[b]{General Motors Global Research \& Development, Warren, MI 48092, United States}

\begin{abstract}
Large language model (LLM) agents that call external tools are increasingly proposed
for manufacturing decision support, yet there is little systematic investigation of
how reliably they orchestrate an existing monitoring pipeline, how they fail, and what
they cost. We present a process-agnostic evaluation harness that wraps a fixed
monitoring model in a tool interface and grades answers by exact match against a
reference computed from that model, so that failures are attributed to the agent's
orchestration rather than to its tools, without a human or model judge. We
demonstrate the harness on an ultrasonic metal welding pipeline that classifies
tool condition (a maintenance decision) and surface condition (a factor in
quality control)
from multi-sensor signals, benchmarking six API-hosted LLM agents from two vendors in
July 2026 on batch-level tasks such as selecting the most worn cycle and finding the
first non-nominal cycle. Correctness on the six tasks ranged from $39\%$ to $100\%$
across models. In a trace-level audit of $72$ attempts, all $15$ failures arose in
orchestration, through incomplete cohort coverage, answers naming the first listed
cycle after all cycles were loaded at once, or fabricated arguments to a zero-argument tool followed by an answer
despite the resulting errors, and several recurred across replicates. The model-group
pattern was stable across a $27$-prompt sensitivity sweep, and enabling extended
thinking produced no accuracy gain on already-saturated tasks. Per-task cost spanned
roughly $16\times$ across models, with input tokens dominating because each turn
resends the growing tool-result history. Evaluating manufacturing decision agents
therefore requires trace-level auditing of orchestration behavior, not only end-task
accuracy. The harness, tasks, and trace logs are publicly released, and the approach
is readily extensible to any manufacturing process with a fixed monitoring model and a
computable decision rule.
\end{abstract}

\cortext[cor1]{Corresponding authors.}

\end{frontmatter}

% Lift the first-page footnote above the journal copyright block that ecrc
% prints at the page foot (the template ships this line commented out). The
% right column of page 1 gets the same adjustment further below; if the first
% page reflows, keep that line in the right column.
% \enlargethispage{-9mm}

%======================================================================
\section{Introduction}
\label{sec:intro}
Large language model (LLM) agents call external tools, keep state across turns, and
combine the returned values into an answer~\citep{yao2023react, wang2024survey}.
Agents that call validated instruments, simulators, and trained predictors have
produced physically validated results in chemistry, materials, and
biology~\citep{boiko2023coscientist, mbran2024chemcrow, szymanski2023alab,
swanson2025virtuallab, ghareeb2026robin}, and manufacturing studies now apply them to
shop-floor coordination~\citep{lim2024llmmultiagent, zhao2025shopfloor}, process
planning~\citep{hoang2026kg, khan2026d2p}, process-parameter optimization in additive
manufacturing~\citep{ajenifujah2025spatter, ajenifujah2026agentam}, literature querying in additive manufacturing~\citep{chandrasekhar2024amgpt}, operations
management~\citep{oberle2026mom}, data analysis~\citep{farid2026agentic}, and defect detection~\citep{halder2026waam,
eslaminia2025fdmbench, eslaminia2026llmadam}. Reviews
and roadmaps name tool integration, verification, and trustworthy decision support as
central open problems~\citep{maghanaki2026llmmanufacturing, lee2025agentic,
leng2026roadmap}, and practitioners cite engineering-grade verification, meaning
testable and replayable agent behavior, as a major adoption
barrier~\citep{edwards2026industry}.

% \enlargethispage{-9mm}% page 1, right column: clear the copyright block
There is limited understanding of how reliably tool-using LLM agents behave in
manufacturing and how they fail.  Many manufacturing-agent studies report one curated demonstration
instead of a systematic comparison across models, replicates, and controlled
conditions~\citep{farid2026agentic, oberle2026mom}.  Two exceptions replicate
comparisons on a physical testbed~\citep{zhao2025shopfloor} or use stratified task
suites with token accounting~\citep{khan2026d2p}.  The closest related study compares
agent configurations over radiography-validated defect classifiers, but it grades
reasoning with a model judge, not a computable reference~\citep{halder2026waam}. General
agent benchmarks target coding, web, and general-knowledge
tasks~\citep{jimenez2024swe, mialon2023gaia, qin2023toolbench, liu2023agentbench}, or
rule-following, stateful tool use, and scientific
analysis~\citep{yao2024taubench, lu2024toolsandbox, chen2024scienceagentbench,
liu2026sciagentarena}, and model-based judging can diverge from expert assessment
where factual correctness matters~\citep{mbran2024chemcrow}. Manufacturing differs
from these settings in three ways. First, the knowledge the decision rests on lives in a
separately validated domain model (a classifier, a regressor, a calibrated sensor
routine), not in the language model itself. Second, decisions are made over batches of parts or
cycles and have exact answers, so a wrong cycle number or count translates directly
into scrap, an escaped defect, or unplanned downtime. Third, decisions must be auditable.
An agent can therefore be wrong while every tool it called was right, and no existing
benchmark measures that case.

To address these gaps, this paper investigates the orchestration reliability and failure modes of tool-using LLM agents for manufacturing decision support.  We target two decisions that are commonly encountered in production: quality (is the workpiece surface in a condition that will produce a high-quality joint?) and maintenance (is the tool worn?).  A technician may ask an agent which cycle in a run is most worn, how many cycles fall below a confidence threshold, or which is the first non-nominal cycle.  These questions are simple once every per-cycle value is available.  Answering them correctly, however, requires the agent to retrieve the value for every required cycle, keep each value bound to its cycle, and apply the stated rule to the complete set.  An answer computed from an incomplete or mis-associated set can still look plausible and well grounded, so the error may be hard to detect.

We propose a process-agnostic evaluation harness for tool-using LLM agents. It wraps a fixed monitoring
model as a set of tools, poses batch-level decision tasks whose answer is a computable
rule over the model's outputs, computes the reference answer directly from the same
model, grades the agent by exact match, and logs every tool call with its returned
value so each failure can be replayed and attributed. Because the reference is the
monitoring model's own output, every error the harness records is an orchestration
error by construction. We use ultrasonic metal welding (UMW) as the testbed for three reasons.  First, UMW is industrially important, most notably for joining tabs and busbars in lithium-ion battery assembly~\citep{lee2014characterization, nong2018robustness}.  Second, it is well studied, with established methods for quality prediction~\citep{wu2022quality, meng2022piel}, tool condition monitoring~\citep{nazir2021tcm, meng2023xfsl, meng2025dg}, and mixed-condition classification~\citep{lu2023online, lu2026sensor}.  Third, our prior work has produced a suite of validated online monitoring models for UMW, which serve directly as the agent's tools in this study~\citep{meng2023xfsl}.

We benchmark six API-hosted models from two vendors, spanning roughly $16\times$ in
per-task cost. Under the same stateful tool interface, correctness on six batch tasks
ranges from $39\%$ to $100\%$ by model. A trace-level audit attributes every failure
to orchestration: incomplete coverage of the cohort, answers naming the first listed
cycle after all cycles were loaded at once, or fabricated tool arguments followed by an answer, several of
them repeating across replicates. The model-group pattern holds across a
$27$-prompt sensitivity sweep, extended thinking yields no gain at ceiling, and input
tokens dominate cost because each turn resends the growing history.

This paper contributes: (i) a process-agnostic evaluation harness and protocol for
tool-using LLM agents over fixed manufacturing monitoring models, with a computable
reference so that failures are graded and attributed without human or model judgment;
(ii) a benchmark of six agents on real UMW quality and maintenance decisions;
(iii) a trace-level catalogue of failure modes that a correctness score hides,
consistent with failure modes reported in other manufacturing agent studies where a
tool fails silently~\citep{surovi2025roles, george2026agentscad}; and (iv) an
operational and sensitivity profile covering latency, tokens, cost, prompt wording,
and extended thinking. The harness, tasks, and
trace logs are released at \repourl{} for extension to newer and open-weight models
and to other manufacturing processes with a fixed monitoring model and a computable
decision rule.

The remainder of the paper is organized as follows. \Cref{sec:methods} introduces
tool-using agents, the UMW monitoring tools, the task set, the models, and the harness
and evaluation protocol. \Cref{sec:results} reports the benchmark, the failure
catalogue, the prompt-sensitivity and extended-thinking results, and the cost
profile. \Cref{sec:discussion} discusses the implications, how the harness extends to
other processes, limitations, and future work, and \cref{sec:conclusion} concludes.

%======================================================================
\section{Methods}
\label{sec:methods}
\Cref{sec:methods-agents} presents the tool-using agent loop.
\Cref{sec:methods-pipeline} introduces the UMW monitoring tools, and
\cref{sec:methods-tasks} describes the task set; these two are the only parts specific to
UMW. \Cref{sec:methods-models} lists the models, \cref{sec:methods-harness} describes
the evaluation harness and protocol, and \cref{sec:methods-metrics} defines the metrics.

\subsection{Tool-using LLM agents}
\label{sec:methods-agents}
\begin{figure*}[t]
  \centering
  \includegraphics[width=\linewidth]{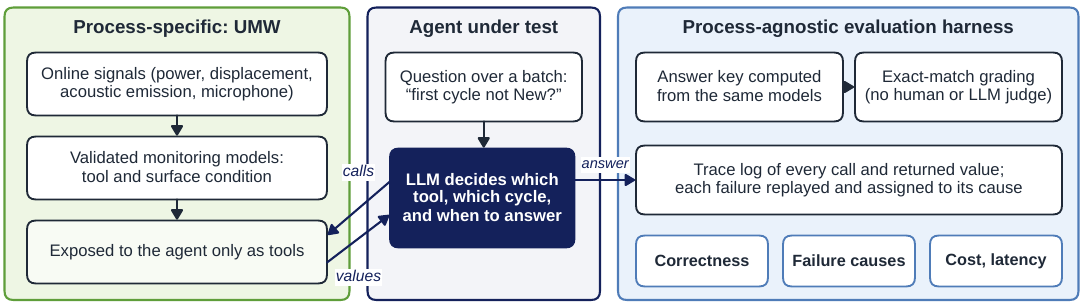}
  \caption{Overview of the framework.  The online signals and the validated monitoring
    models are specific to UMW, and the agent can reach them only through tools (left).
    The LLM under test chooses which tool to call, on which cycle, and when to answer
    (center).  The evaluation harness is process-agnostic.  It grades each
    answer by exact match against a key computed from the same models and logs every
    call, so each failure can be traced to its cause (right).}
  \label{fig:framework}
\end{figure*}

\begin{figure}[t]
  \centering
  \includegraphics[width=\linewidth]{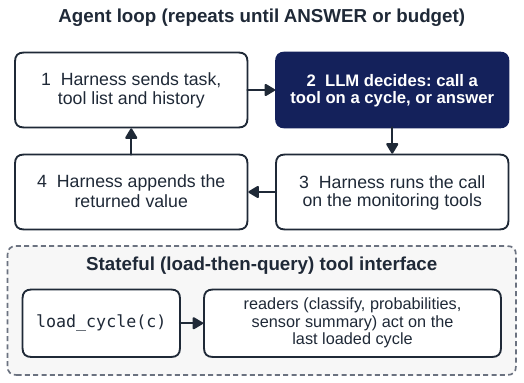}
  \caption{Agent loop and tool interface. Only step~2 is done by the model under test;
    the harness does steps 1, 3 and 4. Under the stateful interface (bottom), the
    readers act on the cycle loaded last, so the agent must track which cycle that is.}
  \label{fig:loop}
\end{figure}

A tool-using LLM agent couples a language model with a set of callable tools and a
control loop~\citep{yao2023react, wang2024survey}. At each turn the model receives the
task, the tool schemas, and the conversation so far, and it either requests one or
more tool calls or commits a final answer. The surrounding software executes each
requested call, appends the result to the conversation, and queries the model again.
In this study (\cref{fig:framework,fig:loop}), the harness performs every execution step
(sending the prompt, running the call, appending the result), and the model under
test performs only the decision step.
Everything the model learns about the process arrives through these tool results; its
own contribution is \emph{orchestration}: choosing which tools to call, on which
items, in what order, when to stop, and how to combine the returned values.

Two properties of this loop matter for manufacturing decisions. First, a batch-level
decision is a rule applied to per-item tool outputs (e.g., the most worn of eight cycles, the
number of cycles below a confidence threshold, the first cycle that is not New), so an
agent can fail while every tool is correct: by skipping an item, attaching a value to
the wrong item, or misapplying the rule. Second, the whole conversation is resent at
every turn, so input tokens, and with them cost and latency, grow with the number of
turns a procedure needs. The tool interface shapes both. Under a stateful
\emph{load-then-query} interface the agent first loads an item into a working context
and then calls tools that act on the most recently loaded item, so it must track that
hidden state across turns; a stateless interface instead names the item in every
call. This study uses the stateful design, which mirrors how the existing UMW monitoring
code is organized.

\subsection{UMW decision-making tools}
\label{sec:methods-pipeline}

A typical UMW system consists of a generator, transducer, booster, horn, and anvil~\citep{meng2023xfsl, meng2022piel}.  During a weld, the horn vibrates at appropriately \SI{20}{\kilo\hertz} against the stationary anvil, and the resulting oscillating shear joins the workpieces clamped under pressure in the solid state.  The horn/anvil tooling and the workpiece surface change over repeated welds, and these changes are difficult to observe directly on the factory floor, so the condition of each cycle is inferred from in-process sensors.  Surface condition is not itself a measure of joint quality, which is defined by the strength of the joint or its class.  Surface condition is an important quality factor because contamination is an uncontrollable disturbance: it alters the welding signals and lowers both joint strength and the rate of successful welds~\citep{lee2014characterization, nong2018robustness, lu2026lmc}.  Classifying surface condition therefore supports quality decisions without being one.

The agent's tools wrap our existing online monitoring pipeline~\citep{meng2023xfsl} without modification.  Each cycle yields four active sensing signals (horn power, horn displacement measured by a linear variable differential transformer (LVDT), acoustic emission, and a microphone), sampled at \SI{200}{\kilo\hertz} for roughly one second.  Each signal is decomposed individually by a discrete wavelet transform (DWT; db4 mother wavelet, thirteen levels), and twelve statistics are computed from the coefficients at each level, giving a $624$-dimensional feature vector per cycle.  Two multilayer perceptrons map this vector to tool condition $\{$New, Worn, Damaged$\}$, a maintenance decision, and surface condition $\{$Clean, Polished, Contaminated$\}$.  The data come from a controlled design of $30$ welds at each of the nine tool and surface combinations ($270$ cycles, $268$ of which load cleanly).  Under leave-one-domain-out (LODO) evaluation, which holds out an entire surface condition when classifying tool condition and vice versa, the tool classifier reaches an accuracy of $0.922$ and the surface classifier $0.568$.  Because the benchmark reference is computed from the classifiers' own outputs, these accuracies limit the physical meaning of an answer but not its grading.  The benchmark tests whether the agent faithfully retrieves and combines the tool outputs, so surface-related results are not claims of physical accuracy.  The pipeline is exposed through ten tools (\cref{tab:tools} in \cref{app:tools}): three file utilities, \ans{load\_cycle}, and six readers that act on the loaded cycle.

\subsection{Task set}
\label{sec:methods-tasks}
Tasks are organized into three tiers by how much the agent must combine. Tier~1 labels
a single cycle, Tier~2 asks for one derived quantity from a single cycle, and Tier~3,
our focus, asks a question about a batch whose answer requires retrieving every
required cycle's value and combining the values under a stated rule. Tier~1 and
Tier~2 baselines saturate for all six models ($18/18$ on each tier: three tasks, one
baseline replicate per model) and are summarized in aggregate. The Tier-3 set has six tasks (\cref{tab:task-set}): five use a common
eight-cycle cohort, and a two-cycle comparison uses the pair (12798, 12869). Three
tasks (argmax-worn, argmax-surface, first-not-new) require ordered selection or
explicit tie handling, two (argmax-lvdt, count-low-conf) require exhaustive iteration
and aggregation, and one (compare) requires cross-cycle comparison. Exact prompts are
in \cref{app:tasks}.

\begin{table*}[t]
\centering
\caption{Advanced (Tier-3) task set. Five batch-level tasks use a common
eight-cycle cohort; the two-cycle comparison (T3-compare) uses a separately
specified pair (12798, 12869). The
reference answer is computable without model-based judgment.}
\label{tab:task-set}
\small
\begin{tabular}{lp{4.4cm}p{3.6cm}l}
\toprule
Task ID & Question & Reference source & Expected \\
\midrule
T3-argmax-worn    & Which cycle is most worn? (highest $P(\mathrm{Worn})$;
ties broken by list order) & Cohort argmax, tool classifier & 12891 \\
T3-compare        & Which sensor differs most between two cycles? (largest
$|\Delta\,\mathrm{mean\_z}|$) & Per-sensor summary difference & Microphone \\
T3-count-low-conf & How many cycles are low-confidence? (predicted-class
probability $<0.99$) & Threshold count, tool classifier & 0 \\
T3-argmax-lvdt    & Which cycle has the largest LVDT signal? (max
\ans{mean\_z}) & Per-sensor summary argmax & 12915 \\
T3-argmax-surface & Which cycle has the highest surface confidence?
(ties broken by list order) & Cohort argmax, surface classifier & 12798 \\
T3-first-not-new  & First cycle in order that is not New (tool prediction
$\neq$ New) & Sequential classifier prediction & 12891 \\
\bottomrule
\end{tabular}
\end{table*}

To separate task difficulty from wording, three Tier-3 tasks (first-not-new,
argmax-worn, argmax-lvdt) and three Tier-1 tasks are each rewritten into nine
variants: three complexity levels (LOW, a bare instruction; MEDIUM, an operational
framing; HIGH, a long narrative with one distractor) crossed with three hand-written
instances, giving $27$ Tier-3 and $27$ Tier-1 prompts. All variants share the cohort
and reference answer and are screened for answer leakage.

\subsection{Models}
\label{sec:methods-models}
\Cref{tab:models} lists the six API-hosted models evaluated in our dated window
(\evalwindow), spanning two vendors, a range of sizes and reasoning configurations, and
roughly $16\times$ in per-task cost. The roster is a comparative sample, not a
leaderboard. Because hosted models change over time, we report the window and the API
identifiers (\cref{app:repro}); swapping in another model is a one-flag change.

\begin{table}[t]
\centering
\caption{Models evaluated. $N$ is replicates per model in the base sweep. o4-mini
emits native reasoning tokens and is always in that mode; the Anthropic models
support optional extended thinking, kept off in the main sweep and switched on for
Opus and Sonnet in the extended-thinking study (\cref{sec:results-thinking}); the two
gpt-4.1 models have no reasoning mode.}
\label{tab:models}
\footnotesize
\begin{tabular}{lllc}
\toprule
Model & Vendor & Reasoning mode & $N$ \\
\midrule
Haiku 4.5    & Anthropic & extended thinking (optional) & 3 \\
Sonnet 4.6   & Anthropic & extended thinking (optional) & 3 \\
Opus 4.7     & Anthropic & extended thinking (optional) & 3 \\
gpt-4.1      & OpenAI    & none                         & 3 \\
gpt-4.1-mini & OpenAI    & none                         & 3 \\
o4-mini      & OpenAI    & native, always on            & 3 \\
\bottomrule
\end{tabular}
\end{table}

\subsection{Evaluation harness and protocol}
\label{sec:methods-harness}
\begin{figure}[t]
  \centering
  \includegraphics[width=\linewidth]{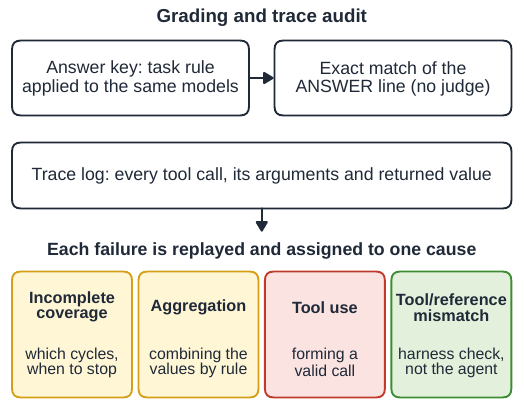}
  \caption{Grading and trace audit. The answer is graded by exact match against a key
    computed from the same monitoring models. Each failure is replayed from the trace
    log and assigned to one cause (\cref{tab:failcat}).}
  \label{fig:audit}
\end{figure}

The harness is the fixed software that hosts the agent and runs the experiment
(\cref{fig:framework}, right; \cref{fig:audit}). It comprises a prompt builder, a vendor-agnostic model client,
a tool dispatcher that wraps the monitoring code, a budget enforcer, and a grader; it
reimplements none of the monitoring code and only calls it and records what returns.
Before the agent runs, the grader computes the reference answer by applying the task
rule to the same monitoring routines the tools expose. The agent then runs the loop of
\cref{sec:methods-agents} on a minimal system prompt (it has tools and an objective;
no domain hints, strategy, or examples) until it emits an \ans{ANSWER:} line or
reaches a budget of $40$ turns, $60$ tool calls, or $4096$ output tokens per call.
Each call and its arguments are logged, and from harness v0.5.5 onward also its
returned value, so a failure can be replayed and attributed by computation.

\emph{Protocol.} Within each comparison the task, interface, budgets, and harness are
fixed and one factor varies: the model (base sweep: six Tier-3 tasks $\times$ six
models $\times$ three replicates $= 108$ attempts), the prompt wording (sensitivity
sweep: $27$ prompts $\times$ six models $\times$ three replicates $= 486$ attempts per
tier), or the thinking configuration (extended thinking on or off for Opus and
Sonnet). A trace audit re-runs the Tier-3 tasks with per-call result logging and
replays every failure on the four rule-following and iteration families ($72$
attempts). Sampling parameters are left at provider defaults, so replicate-to-replicate
variation reflects each provider's stochastic decoding. The harness evolved through
revisions that each removed a specific evaluation confound
(\cref{tab:harness-evolution} in \cref{app:repro}).

\emph{Requirements for applying the harness to another system.} Only the
process-specific layers in \cref{fig:framework} change. A target system must provide
five things: (i) a fixed monitoring model or set of routines, versioned and
deterministic, so that the same item always returns the same value; (ii) programmatic
access to those routines per item, so that each can be wrapped as a tool with a name,
arguments, and a serializable return value; (iii) addressable items (cycles, parts, or
batches) with stable identifiers; (iv) decision tasks whose answer is a stated,
computable rule over per-item outputs (selection, counting, first-match, comparison, or
aggregation), with ties and thresholds specified; and (v) a reference function that
applies that rule directly to the same routines, with an answer format (an identifier,
an integer, or a label) that admits exact matching. The model client, budgets, grader,
logging, and trace replay are unchanged. Physical ground truth is optional: without it
the harness measures orchestration error only, and with it the same protocol can also
measure the monitoring model's own error. Open-ended advisory questions without an
objective key are out of scope, because grading them requires a human or model judge.

\subsection{Metrics}
\label{sec:methods-metrics}
Correctness is exact match between the value on the terminal \ans{ANSWER:} line and
the reference; when a cycle number is expected, only that number counts, and no
model-based judgment or fuzzy matching is used. Per attempt we log tool calls, turns,
wall-clock latency, input, output, and reasoning tokens, and the ordered tool-call
sequence with returned values; cost uses provider list prices in the evaluation
window. In the trace audit, each failure is assigned to one category by replaying its
log, and each category is tied to the agent decision that caused it:
\emph{incomplete coverage}, a decision about which items to query or when to stop (a
cycle that decides the answer was never queried); \emph{aggregation}, a decision about
how to combine the values (every decisive value was retrieved but the rule was
misapplied); and \emph{tool use}, a decision about how to form a call (an invalid
invocation, such as fabricated arguments). A fourth category, \emph{tool/reference
mismatch} (a valid returned value differs from the value the reference used), lies on
the tool side rather than with the agent and serves as a consistency check of the
harness.

%======================================================================
\section{Results}
\label{sec:results}

\subsection{Correctness on batch-level decisions}
\label{sec:results-base}
\Cref{tab:overall-scores} reports the base sweep ($108$ attempts). Haiku, Sonnet, and
Opus answered all $54$ of their attempts correctly and o4-mini $17$ of $18$;
gpt-4.1-mini scored $12/18$ and gpt-4.1 $7/18$ ($39\%$). Grouped, the four models
whose families offer a reasoning or thinking mode scored $71/72$ ($99\%$) and the two
GPT-4.1 models $19/36$ ($53\%$). Extended thinking was off for the Anthropic models in
this sweep, so the grouping is by model family rather than by reasoning use. Per task
(\cref{tab:per-task}), counting saturated ($18/18$) and the two-cycle comparison
nearly so ($17/18$), while the rule-following and iteration tasks held the failures
($13$ to $15$ of $18$).

\begin{table}[t]
\centering
\caption{Base Tier-3 sweep. Each cell is correct-of-six in that replicate; Total is
$/18$. The upper block lists the models whose families offer a reasoning or thinking
mode (thinking off for the Anthropic models in this sweep).}
\label{tab:overall-scores}
\small
\begin{tabular}{lcccc}
\toprule
Model & Run 1 & Run 2 & Run 3 & Total \\
\midrule
Haiku 4.5  & 6/6 & 6/6 & 6/6 & 18/18 \\
Sonnet 4.6 & 6/6 & 6/6 & 6/6 & 18/18 \\
Opus 4.7   & 6/6 & 6/6 & 6/6 & 18/18 \\
o4-mini    & 6/6 & 5/6 & 6/6 & 17/18 \\
\midrule
gpt-4.1-mini & 2/6 & 5/6 & 5/6 & 12/18 \\
gpt-4.1      & 2/6 & 2/6 & 3/6 & 7/18 \\
\bottomrule
\end{tabular}
\end{table}

\begin{table}[t]
\centering
\caption{Per-task correctness across six models and three replicates ($18$ attempts
per task).}
\label{tab:per-task}
\small
\begin{tabular}{lc}
\toprule
Task & Correct (of 18) \\
\midrule
T3-count-low-conf & 18 \\
T3-compare        & 17 \\
T3-argmax-lvdt    & 15 \\
T3-first-not-new  & 14 \\
T3-argmax-worn    & 13 \\
T3-argmax-surface & 13 \\
\bottomrule
\end{tabular}
\end{table}

\subsection{Trace-level failure catalogue}
\label{sec:results-catalogue}
The audit re-ran the Tier-3 suite under harness v0.5.5, which changes only logging,
and replayed every failure on the four rule-following and iteration families
($4$ tasks $\times$ $6$ models $\times$ $3$ replicates $= 72$ attempts). Of these,
$57$ succeeded and $15$ failed, close to the base sweep's $55/72$ on the same
families. The count task (no failures) and the two-cycle comparison (no cohort
iteration) were excluded. The $15$ failures comprise $10$ incomplete-coverage, $4$
aggregation, and $1$ tool-use failure, and the two GPT-4.1 models account for $13$
(\cref{tab:failcat}). As a consistency check, no valid returned value differed from
the value used by the reference, confirming that each failure is attributable to
orchestration. Two modes recur.

\begin{figure*}[t]
\centering
\includegraphics[width=\linewidth]{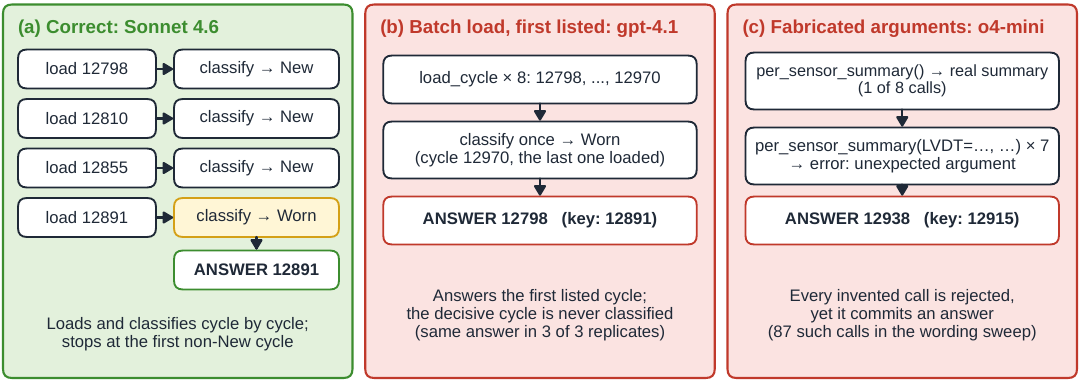}
\caption{Three audit-sweep traces under the stateful interface. (a) Correct:
Sonnet~4.6 on T3-first-not-new loads and classifies each cycle in order and stops at
the first non-New cycle ($12$ tool calls over $11$ turns; four file look-ups not shown). (b) First-listed answer after a batch load, with incomplete coverage: gpt-4.1 on
T3-first-not-new loads all eight cycles, classifies only the last, and answers the
first listed cycle; the same answer occurred in all three replicates. (c) Fabricated
tool arguments: o4-mini on T3-argmax-lvdt passes invented sensor values to the
zero-argument summary tool on seven of its eight calls; each call is rejected, yet it
commits an answer. This pattern recurs on $87$ calls in the prompt-sensitivity sweep
and was not observed in the tested Anthropic runs.}
\label{fig:traces}
\end{figure*}

\emph{First-listed answers after a batch load.} Nine of the $15$ failures answered
the first listed cycle, all by the two GPT-4.1 models, and in all nine the agent had
issued eight load calls in a row before reading any value. Because the tools keep only
the last loaded cycle, the readings that followed returned that cycle's values. In the four aggregation failures the agent had retrieved the winning value and
still returned the first cycle; in five coverage failures it never obtained the
decisive value. On T3-first-not-new, for example, gpt-4.1 answered $12798$ instead of
$12891$ in all three replicates: it loaded all eight cycles, classified only the last,
and committed the first (\cref{fig:traces}b), using $9$ of $60$ tool calls and $3$ of
$40$ turns, so the failure is not a budget artifact. Because the decisive value was
never obtained, the audit counts this case as incomplete coverage. The first-listed
answer is therefore a behavioral signature that cuts across the attribution categories.
A follow-up run in October 2026 ties it to the interface: with the API's parallel tool
calls disabled, so that each turn carries a single call, gpt-4.1 answered $60$ of $60$
base-sweep attempts correctly over $10$ replicates, against $25$ of $60$ with the default
setting.

\emph{Fabricated tool arguments.} On T3-argmax-lvdt, o4-mini called the zero-argument
\ans{per\_sensor\_summary} tool with a constructed dictionary of sensor values on
seven of its eight summary calls. The harness rejected each such call with an
explicit error (\cref{fig:traces}c), so a real summary was returned for
only one cycle, yet the model committed to a specific cycle. The same argument
pattern appears in o4-mini's single base-sweep failure, and the prompt-sensitivity
sweep reproduces it on $87$ calls across all nine argmax-lvdt variants.

\begin{table*}[t]
\centering
\caption{Failure catalogue over the four trace-audited rule-following and iteration
families ($72$ attempts, $15$ failures). The two GPT-4.1 models account for $13$ of
the $15$. Categories are defined in \cref{sec:methods-metrics}.}
\label{tab:failcat}
\small
\begin{tabular}{lcp{4.1cm}p{8.2cm}}
\toprule
Failure category & Count & Agent decision at fault & Observed mechanism \\
\midrule
Incomplete coverage & $10$ & Which cycles to query; when to stop & Decisive cycle never
queried; answer computed over an incomplete set ($5$ of the $10$ answered the first
listed cycle). \\
Aggregation & $4$ & Combining values under the rule & Winning value retrieved, but the
first listed cycle returned. \\
Tool use & $1$ & Forming a valid tool call & Invented arguments to a zero-argument tool;
answer committed despite the errors (o4-mini). \\
\midrule
Tool/reference mismatch & $0$ & None: tool side (consistency check) & Valid returned
value differed from the value used by the reference; never observed. \\
\bottomrule
\end{tabular}
\end{table*}

\subsection{Prompt sensitivity}
\label{sec:results-complexity}
\begin{figure*}[!tb]
\centering
\includegraphics[width=\linewidth]{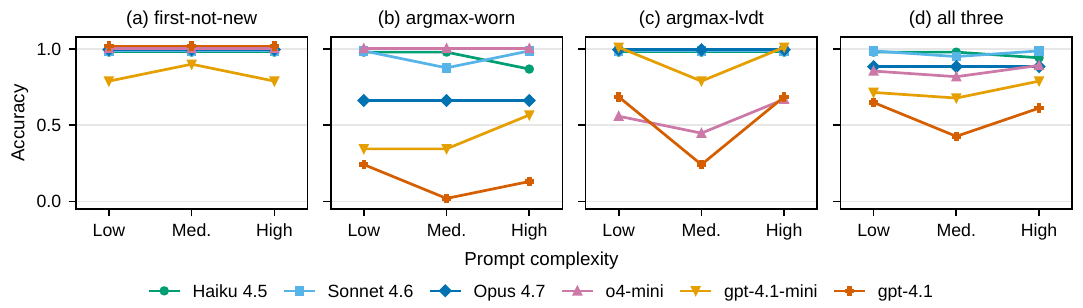}
\caption{Tier-3 accuracy versus prompt complexity (Low, Medium, High) on the
prompt-sensitivity sweep, per model: (a) T3-first-not-new, (b) T3-argmax-worn,
(c) T3-argmax-lvdt, and (d) all three tasks pooled. Each point is $9$ attempts per
model in (a) to (c) and $27$ in (d); overlapping lines at $1.0$ hide models that
answered every variant correctly.}
\label{fig:degradation}
\end{figure*}

Every model answered every Tier-1 variant correctly ($486/486$). On the Tier-3
variants (\cref{tab:complexity-tier3}), the model-group pattern of
the base sweep held: the four models whose families offer a reasoning or thinking
mode scored $301/324$ ($92.9\%$) and the two GPT-4.1 models $102/162$ ($63.0\%$).
Accuracy was flat or non-monotonic in complexity level rather than decreasing
(\cref{fig:degradation}), and each model's errors concentrated on one task. The sweep
reproduced the catalogued modes at scale: gpt-4.1-mini answered the first listed cycle on $16$ of $27$
argmax-worn variants, and o4-mini fabricated arguments on $87$ argmax-lvdt calls.

T3-argmax-worn is numerically fragile: five of its eight cycles saturate at
$P(\mathrm{Worn}) \approx 1.0$, so the answer rests on the tie-break and on residuals
of order $10^{-9}$. All errors by the Anthropic models occurred on this task: Opus
($9$ of $27$), Haiku, and Sonnet (one each) demoted the tie-break winner after reading
a $\sim\!2.6\times10^{-9}$ residual. Results on this task therefore mix rule-following
with numerical tie handling.

\begin{table}[!htb]
\centering
\caption{Tier-3 prompt-sensitivity sweep ($27$ prompts $\times$ $6$ models $\times$ $3$
replicates $= 486$ attempts). Errors among the Anthropic models are confined to the
numerically fragile T3-argmax-worn task.}
\label{tab:complexity-tier3}
\small
\begin{tabular}{lcc}
\toprule
Model & Accuracy & Correct / attempts \\
\midrule
Haiku 4.5    & $98.8\%$ & $80/81$ \\
Sonnet 4.6   & $98.8\%$ & $80/81$ \\
Opus 4.7     & $88.9\%$ & $72/81$ \\
o4-mini      & $85.2\%$ & $69/81$ \\
gpt-4.1-mini & $71.6\%$ & $58/81$ \\
gpt-4.1      & $54.3\%$ & $44/81$ \\
\bottomrule
\end{tabular}
\end{table}

\subsection{Extended thinking}
\label{sec:results-thinking}
Enabling extended thinking on Opus and Sonnet left correctness unchanged on tasks they
already solved (\cref{tab:thinking}); this study used the three-task Tier-3 set of
harness v0.4 (argmax-worn, compare, count-low-conf). Across all twelve logged thinking-enabled Opus
executions (six Tier-3 attempts over two replicates, plus three Tier-1 and three
Tier-2 tasks), adaptive thinking allocated zero reasoning tokens and matched the
baseline's correctness and tool-call counts. Sonnet with a fixed $8192$-token budget
spent $200$ to $600$ thinking tokens per Tier-3 task without changing correctness, and
on cold-start runs its argmax tool calls rose from $17$ to $24$ through additional
per-cycle verification.

\begin{table*}[!tb]
\centering
\caption{Extended-thinking experiment. Correctness is unchanged from baseline; token
allocation differs by system. Warm denotes the argmax task executed within the
full nine-task sweep; cold denotes a standalone fresh-invocation run of the
three advanced tasks. Each task is an independent conversation in both cases,
so the labels describe run context, not shared model state.}
\label{tab:thinking}
\small
\begin{tabular}{lccc}
\toprule
Model + config & Correctness & Thinking tok (observed) & Tool calls (argmax) \\
\midrule
Opus, no thinking    & 6/6 & n/a & 11 \\
Opus, adaptive-high  & 6/6 & 0        & 11 (unchanged) \\
Sonnet, no thinking  & 6/6 & n/a & 17 \\
Sonnet, budget-8192  & 6/6 & 200 to 600 & 17 warm / 24 cold \\
\bottomrule
\end{tabular}
\end{table*}

\subsection{Operational cost}
\label{sec:results-cost}
\Cref{tab:operational} reports per-task latency, tokens, tool calls, and cost on the
prompt-sensitivity sweep. Per-task cost spanned roughly $16\times$ (\$0.02 to
\$0.32). Input tokens exceeded output tokens by one to two orders of magnitude for
every model (for example, $75{,}608$ versus $1{,}583$ for Haiku), because each turn
resends the growing history. Reasoning tokens made up about $69\%$ of o4-mini's
output, billed at output rates. The full $486$-attempt Tier-3 sweep cost \$61.99, of
which Opus accounted for $42\%$.

\begin{table*}[!tb]
\centering
\caption{Operational profile on the Tier-3 prompt-sensitivity sweep: per-task means over
$81$ attempts per model. Cost at provider list prices, \evalwindow.}
\label{tab:operational}
\small
\begin{tabular}{lrrrrr}
\toprule
Model & Latency & Input tok & Output tok & Tool calls & Cost/task \\
\midrule
Haiku 4.5    & $24.2$s & $75{,}608$ & $1{,}583$ & $21.0$ & \$0.08 \\
Sonnet 4.6   & $41.0$s & $70{,}128$ & $1{,}586$ & $19.9$ & \$0.23 \\
Opus 4.7     & $38.6$s & $58{,}067$ & $1{,}382$ & $14.6$ & \$0.32 \\
o4-mini      & $28.3$s & $39{,}756$ & $2{,}146$ & $15.8$ & \$0.05 \\
gpt-4.1-mini & $25.5$s & $48{,}817$ & $863$     & $32.3$ & \$0.02 \\
gpt-4.1      & $12.3$s & $22{,}506$ & $460$     & $17.8$ & \$0.05 \\
\bottomrule
\end{tabular}
\end{table*}

%======================================================================
\section{Discussion}
\label{sec:discussion}

\subsection{Reliability is limited by orchestration, not by the monitoring model}
\label{sec:disc-orchestration}
Every audited failure arose while the agent gathered and combined values that the
tools returned correctly.  Manufacturing practice concentrates its verification effort
on the monitoring model, but validating that model does not validate the agent built
on it.  A checked classifier can still yield a wrong batch decision if the agent skips
a cycle, attaches a value to the wrong cycle, or makes an invalid call.  The
correctness spread in \cref{tab:overall-scores} is also not a capability ranking.  It
was measured under one stateful interface, with model family, scale, training, and
reasoning support left uncontrolled.  The traces suggest that much of the gap comes
from tracking the loaded cycle across a long trajectory rather than from the
arithmetic.  Testing this hypothesis
directly, through interventions on the tool interface, is ongoing work
(\cref{sec:disc-future}).

\subsection{What a correctness score hides}
\label{sec:disc-modes}
The two catalogued modes carry opposite deployment risks.  A first-listed answer
returns a plausible but wrong cycle that a technician has no reason to question.
Argument fabrication produces an answer that no valid tool return supports, although
the errors are visible in the log.  o4-mini's high score hides the second mode
entirely, and neither mode can be seen without a full tool-call trace.
Manufacturing-agent evaluations should therefore report failure modes and traces
alongside accuracy, and deployed agents should log each tool call with its returned
value so that a questionable decision can be replayed.

\subsection{Cost, interface, and reasoning configuration}
\label{sec:disc-cost}
Model choice is a first-order cost decision: for the same task, cost differed by up to
$16\times$ across models.  Because input tokens dominate and grow with the number of
turns, the tool interface affects cost on the same order.  An interface that needs
fewer, more self-contained calls per cycle shortens the history that every turn
rereads.  Extended thinking added tokens without adding accuracy here, but every task
run with thinking enabled was already solved without it.  Evaluations of reasoning
modes should include tasks below the ceiling, where a benefit could appear.

\subsection{Generality of the harness}
\label{sec:disc-extension}
The harness separates the process-specific parts (the tool wrappers and the task
rules) from the rest (the agent loop, budgets, grading, logging, and trace replay).
Any process with a fixed monitoring model and decisions that reduce to a computable
rule over its outputs can be evaluated by supplying those two parts and a reference
function.  Examples include tool-wear monitoring in machining and defect
classification in additive manufacturing~\citep{ajenifujah2024spatterml}.  Every task here could also be answered by
a short deterministic script, and that script is how the reference is computed.  The
agent is useful because it accepts requests phrased in natural language without a
pre-written rule for each one.  The harness measures how much reliability is lost and
how much cost is added in exchange for that flexibility, with the script setting both
the accuracy reference and the cost floor.

\subsection{Limitations}
\label{sec:disc-limitations}
All results come from one process, one classification pipeline, one tool interface,
and two vendors, with three replicates for the base and prompt-sensitivity
conditions.  The tasks are objectively gradable retrieval and aggregation rules
rather than open-ended engineering reasoning, and T3-argmax-worn is numerically
fragile.  The grouping of models by reasoning support is observational.  Correctness
is defined relative to the fixed monitoring outputs, not physical weld ground truth.
An agent graded correct has faithfully retrieved and combined those outputs, which is
a separate question from whether they are physically right.  By the same design, the
harness isolates orchestration error and cannot detect an error in the monitoring
model itself.  Finally, hosted models change over time, so the results describe the
configurations served in our window (\evalwindow), not permanent properties of the
model families.

\subsection{Future work}
\label{sec:disc-future}
First, the stateful load-then-query design could be replaced with tools that name the
cycle in every call, and the batch scaled toward production size.  This would test
how far the first-listed and coverage failures are interface effects, as the
follow-up result for gpt-4.1 in \cref{sec:results-catalogue}
suggests; this work is under way. Second, the harness could be applied to a regression
pipeline graded against destructive measurements, such as joint-strength
prediction~\citep{meng2022piel}, so that monitoring-model error and orchestration
error can be separated.  Third, coverage could be broadened to open-weight models that a
plant can run on its own hardware, newer hosted models, other processes, and tasks with partially computable keys.
Fourth, it remains open whether multi-agent designs with explicit memory, or
deterministic gathering paired with an LLM aggregator, meet a stated reliability and
cost target better than a single agent.

%======================================================================
\section{Conclusion}
\label{sec:conclusion}
We presented a process-agnostic harness for evaluating tool-using LLM agents that
make decisions over fixed manufacturing monitoring models, and demonstrated it on
ultrasonic metal welding quality and maintenance decisions.  Under a stateful
load-then-query interface, six API-hosted agents ranged from $39\%$ to $100\%$
correct on batch-level decisions that a short deterministic script answers exactly,
and this pattern held across $27$ prompt variants.  Trace replay attributed every
audited failure to orchestration rather than to the monitoring model: incomplete
coverage, first-listed answers after loading all cycles at once, or fabricated tool
arguments.  These failures recurred across replicates rather than at random.  The
first-listed answer carries the greater risk, because it returns a plausible cycle
that a technician has no reason to question.  The interface appears to be a major
lever.  Disabling parallel tool calls raised gpt-4.1 from $25$ to $60$ of $60$
correct, and because input tokens dominate a per-task cost that spanned $16\times$
across models, the interface also shapes cost.

These results establish that, for manufacturing decision agents, validating
the monitoring model does not validate the agent built on it: reliability is
determined by how the agent gathers and combines model outputs, and its most
consequential errors are plausible answers that accuracy alone cannot reveal.  The
harness gives practitioners a process-agnostic way to measure this before
deployment, quantifying the reliability and cost traded for natural-language
flexibility against a deterministic reference.  Manufacturing decision agents
should therefore be qualified with trace-level audits as well as accuracy, and
deployed agents should log every tool call with its returned value.

% \section*{CRediT authorship contribution statement}
% \todo{Author names withheld for review.} Conceptualization; Methodology; Software;
% Investigation; Formal analysis; Data curation; Writing -- original draft; Writing --
% review \& editing; Visualization; Supervision.
%
% \section*{Declaration of competing interest}
% The authors declare the following interests that may be considered competing: one or
% more authors are employed by General Motors, a collaborator on this project.
%
% \section*{Data availability}
% The evaluation harness, task definitions, analysis scripts, and anonymized agent
% trace logs are publicly available at \repourl. The underlying ultrasonic metal welding sensor
% dataset is available from the authors upon reasonable request.

%======================================================================

\appendix
% Restart table and figure numbering in each appendix (Table A.1, B.1, C.1).
\makeatletter
\@addtoreset{table}{section}
\@addtoreset{figure}{section}
\makeatother
\section{Tool inventory}
\label{app:tools}
\Cref{tab:tools} lists the ten tools exposed to the agent under the load-then-query
interface. Each wraps an existing UMW monitoring routine; the harness reimplements
nothing. The agent receives no guidance on which tools to use or in what order.

\begin{table*}[!tb]
\centering
\caption{Tools exposed to the agent (load-then-query interface). ``Ctx'' = operates on
the most-recently-loaded cycle.}
\label{tab:tools}
\footnotesize
\begin{tabular}{p{3.4cm}p{3.0cm}p{2.0cm}p{6.0cm}}
\toprule
Tool & Arguments & Ctx & Returns \\
\midrule
\texttt{list\_dataset\_files} & pattern & n/a & matching file paths \\
\texttt{read\_file\_head} & path, n\_lines & n/a & first lines of a file \\
\texttt{grep\_text} & pattern, path & n/a & matching lines \\
\texttt{load\_cycle} & cycle\_path & sets ctx & loads a cycle into working state \\
\texttt{classify} & classifier & loaded & prediction + confidence \\
\texttt{per\_class\_probabilities} & classifier & loaded & full class distribution \\
\texttt{per\_sensor\_summary} & none & loaded & per-sensor \texttt{mean\_z} etc. \\
\texttt{top\_features} & classifier, top\_k & loaded & top distinguishing features \\
\texttt{calibration} & classifier & loaded & LODO calibration metadata \\
\texttt{signal\_slice} & sensor, times & loaded & raw signal slice \\
\bottomrule
\end{tabular}
\end{table*}

\section{Task catalogue}
\label{app:tasks}
Every task is delivered as a single user turn, on a minimal system prompt that states
only that the agent has tools and an objective and must end with a line
\ans{ANSWER: <value>}. The system prompt contains no domain-specific hints,
strategy, or worked examples; task-specific framing appears only in the user
prompt according to the experimental variant, and tool schemas are supplied
through the API. \Cref{tab:task-catalogue}
gives the exact baseline (V0) question text, cohort, and reference-answer rule for the six
advanced tasks; five batch-level tasks share the fixed cohort $\{$12798, 12810,
12855, 12891, 12915, 12923, 12938, 12970$\}$, and the two-cycle comparison
(T3-compare) uses the separately specified pair (12798, 12869). For the prompt-sensitivity study, each selected task is written into nine variants (three
complexity levels, LOW a bare instruction, MEDIUM an operational framing, HIGH a
long narrative with a distractor, crossed with three hand-written instances),
yielding $27$ prompts across the three tasks; all variants share the cohort and
reference answer, and only the wording changes.

\begin{table*}[!tb]
\centering
\caption{The six advanced (Tier-3) tasks. Full prompts reproduced verbatim (dataset
path elided).}
\label{tab:task-catalogue}
\footnotesize
\begin{tabular}{p{2.4cm}p{2.0cm}p{2.0cm}p{8.4cm}}
\toprule
Task id & Per-cycle value & Aggregation & Baseline (V0) prompt \\
\midrule
T3-argmax-worn & $P(\mathrm{Worn})$ (tool) & argmax, first-listed tie-break &
``\ldots which one has the highest tool-classifier probability for the Worn class?
Cycles to consider: \ldots. If multiple cycles tie at the maximum probability \ldots
return the first cycle in the order listed above. Provide just the cycle number
\ldots'' \\
T3-argmax-surface & surface confidence & argmax, first-listed tie-break &
``\ldots which one has the highest surface-classifier confidence for its predicted
class? \ldots If multiple cycles tie \ldots return the first cycle in the order listed
\ldots'' \\
T3-argmax-lvdt & LVDT \texttt{mean\_z} & argmax &
``\ldots which one has the highest LVDT sensor mean\_z value? Use the
per\_sensor\_summary tool \ldots Cycles to consider: \ldots'' \\
T3-first-not-new & tool prediction & first-match &
``\ldots in the listed order, what is the first cycle whose tool-classifier
prediction is NOT `New'? Cycles to consider: \ldots'' \\
T3-count-low-conf & $\mathbb{1}[\text{conf}<0.99]$ & count &
``\ldots how many of these cycles have a tool-classifier confidence strictly below
0.99 for their predicted class? \ldots Provide just an integer \ldots'' \\
T3-compare & $|\Delta\,\texttt{mean\_z}|$ & argmax over sensors &
``Compare cycles 12798 and 12869 \ldots Which sensor has the largest absolute
difference in mean\_z between the two cycles? \ldots name that sensor.'' \\
\bottomrule
\end{tabular}
\end{table*}

\section{Reproducibility}
\label{app:repro}
Main benchmark, trace-audit, and prompt-sensitivity runs use $40$ turns, $60$
tool calls, and $4096$ output tokens per model call. Sampling parameters
(temperature, top-p, seed) are never set explicitly: every request uses the
provider's default sampling configuration, and the o-series endpoint accepts no
temperature parameter, so run-to-run variation across replicates reflects each
provider's default stochastic decoding. The extended-thinking
experiment uses the provider-specific thinking configurations reported in
\cref{tab:thinking}: Opus in adaptive thinking at effort ``high,'' Sonnet with a
fixed $8192$-token thinking budget, and, for thinking-enabled runs only, an
output cap raised to $16384$ tokens with the interleaved-thinking beta header.
Grading is
exact match on the terminal \ans{ANSWER:} line; extended thinking is off in the main
sweep. We evaluate whatever hosted versions the providers served for the chosen
identifiers during our evaluation window
(\evalwindow); the API model identifiers used were \ans{claude-haiku-4-5},
\ans{claude-sonnet-4-6}, \ans{claude-opus-4-7}, \ans{o4-mini}, \ans{gpt-4.1},
and \ans{gpt-4.1-mini}, and they ship with the code; the harness is
model-agnostic, so a new model is one \texttt{--model} flag. Costs use list prices in USD
per million input and output tokens: Haiku 4.5 $1$ and $5$, Sonnet 4.6 $3$ and $15$, Opus
4.7 $5$ and $25$, o4-mini $1.10$ and $4.40$, gpt-4.1-mini $0.40$ and $1.60$, and gpt-4.1
$2$ and $8$. \Cref{tab:harness-evolution}
records the harness decisions, each of which removed a specific evaluation
confound or improved traceability, so every result can be defended against the
confound it would otherwise carry.

\begin{table*}[!tb]
\centering
\caption{Harness evolution: each revision removed a specific evaluation confound
or improved traceability.}
\label{tab:harness-evolution}
\small
\begin{tabular}{lp{3.9cm}p{5.9cm}}
\toprule
Version & Change & Purpose \\
\midrule
v0 $\rightarrow$ v0.1 &
\textbf{Budget headroom:} turns $15{\to}40$, tool calls $25{\to}60$, tokens
$2048{\to}4096$ &
Capability failure confounded with budget exhaustion; argmax tasks need $11$ to $24$
tool calls. \\[2pt]
v0.1 $\rightarrow$ v0.2 &
\textbf{Tie-break rule:} explicit tie-breaking added to argmax wording &
Grading unfairness: a defensible tie-break answer scored wrong; separates
rule-following from rule-choice. \\[2pt]
v0.2 $\rightarrow$ v0.3 &
\textbf{API contracts:} vendor-appropriate (\ans{max\_completion\_tokens}; no
\ans{temperature} for reasoning models) &
OpenAI o-series silently returned empty responses to Anthropic-style payloads. \\[2pt]
v0.3 $\rightarrow$ v0.4 &
\textbf{Extended thinking:} Anthropic support (adaptive for Opus, fixed
\ans{budget\_tokens} for Sonnet) &
Enabled the extended-thinking test (\cref{sec:results-thinking}). \\[2pt]
v0.4 $\rightarrow$ v0.5 &
\textbf{Task expansion:} from three to six advanced tasks &
Per-task quirks; first-not-new was needed to distinguish first-listed answers from genuine iteration. \\[2pt]
v0.5 $\rightarrow$ v0.5.3 &
\textbf{Prompt-sensitivity sets:} Tier-1 and Tier-3 prompt sets; \ans{argmax-lvdt}
disambiguated to the \ans{mean\_z} statistic &
Statistic ambiguity (\ans{mean\_z} vs.\ \ans{max\_z}) that scored a defensible reading
wrong; enabled the prompt-sensitivity sweep. \\[2pt]
v0.5.3 $\rightarrow$ v0.5.5 &
\textbf{Per-call result logging:} each tool call's returned value stored in the
run log &
Reliance on manual failure attribution; per-call outputs enable trace-based
computational attribution (\cref{sec:results-catalogue}, 72-attempt re-run of
the four rule-following and iteration families). \\
\bottomrule
\end{tabular}
\end{table*}

% \FloatBarrier   % removed: forced a float-only page and pushed references to a new page
%% References
\bibliographystyle{elsarticle-num}
\bibliography{reference}

\end{document}